\documentclass{article}
\usepackage{graphicx}
\usepackage{amsmath}
\usepackage{algorithm}
\usepackage{algpseudocode}
\PassOptionsToPackage{numbers,square}{natbib}
\usepackage{hyperref}

\usepackage[dblblindworkshop, final]{neurips_2025}
\workshoptitle{Embodied World Models for Decision Making}

\usepackage[utf8]{inputenc} 
\usepackage[T1]{fontenc}    
\usepackage{hyperref}       
\usepackage{url}            
\usepackage{booktabs}       
\usepackage{amsfonts}       
\usepackage{nicefrac}       
\usepackage{microtype}      
\usepackage{xcolor}         

\title{Plan Verification for LLM-Based Embodied Task Completion Agents}

\author{
  Ananth Hariharan, Vardhan Dongre, Dilek Hakkani-Tür, Gokhan Tur\\
  University of Illinois Urbana-Champaign\\
  \texttt{\{ahari8, vdongre2, dilek, gokhan\}@illinois.edu} 
}

\begin{document}

\maketitle


\begin{abstract}
Large language model (LLM) based task plans and corresponding human demonstrations for embodied AI may be noisy, with unnecessary actions, redundant navigation, and logical errors that reduce policy quality. We propose an iterative verification framework in which a Judge LLM critiques action sequences and a Planner LLM applies the revisions, yielding progressively cleaner and more spatially coherent trajectories. Unlike rule-based approaches, our method relies on natural language prompting, enabling broad generalization across error types including irrelevant actions, contradictions, and missing steps. On a set of manually annotated actions from the TEACh embodied AI dataset, our framework achieves up to 90\% recall and 100\% precision across four state-of-the-art LLMs (GPT o4-mini, DeepSeek-R1, Gemini 2.5, LLaMA 4 Scout). The refinement loop converges quickly, with 96.5\% of sequences requiring at most three iterations, while improving both temporal efficiency and spatial action organization. Crucially, the method preserves human error-recovery patterns rather than collapsing them, supporting future work on robust corrective behavior. By establishing plan verification as a reliable LLM capability for spatial planning and action refinement, we provide a scalable path to higher-quality training data for imitation learning in embodied AI. \footnote{Code and models: \url{https://github.com/AnanthHariharan/Task-Agents}}
\end{abstract}

\begin{figure}[t]
\begin{center}
\includegraphics[width=12cm]{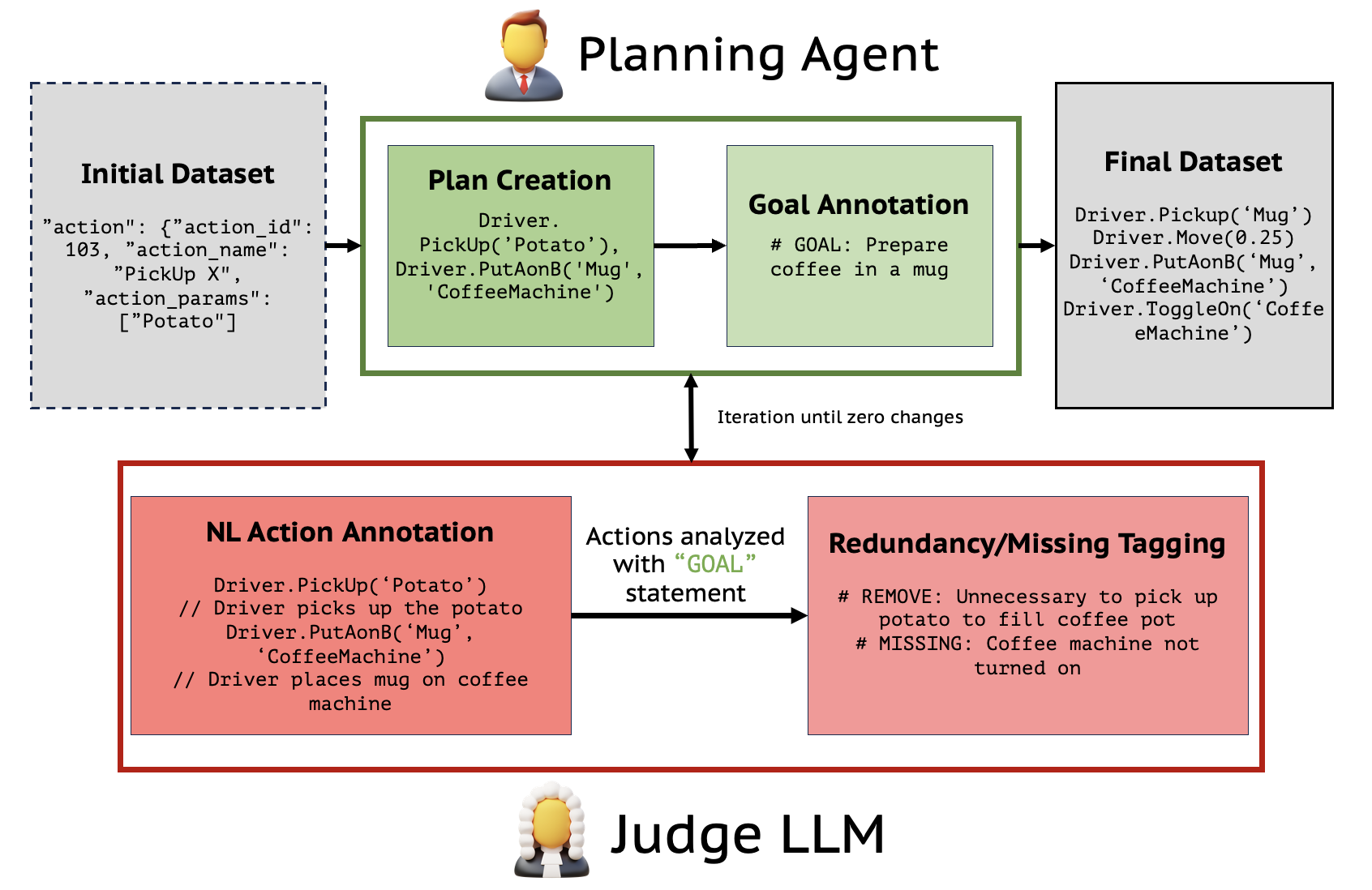} \hfill
  \caption {Diagram of Planning Agent and Judge LLM Interaction Process for Plan Verification} 
\label{fig:multiagent-workflow}
\end{center}
\end{figure}


\section{Introduction}

Recent advances in large language models (LLMs) have enabled sophisticated multi-agent systems for handling task plans in embodied AI. Embodied agents deployed in real-world environments are often tasked with executing extended sequences of actions in pursuit of high-level goals. For such agents to be effective, their plans must be accurate in terms of task completion, efficient in their use of time and actions, and safe in avoiding undesirable or unintended consequences. With the advent of LLMs, there has been growing interest in using these systems to generate and reason about structured plans from natural language input. Recent work has demonstrated that LLMs can decompose high-level goals into executable sub-goals and, in multi-agent setups, evaluate and revise each other’s outputs through structured dialogue (\citep{shinn2023reflexion}, ~\citep{wu2023autogen}).

Despite this progress, LLM-generated plans have been shown to be not perfect~\cite[among others]{padmakumar-etal-2023-multimodal}, and need to be verified. Furthermore, the quality of the datasets used to train planning for embodied agents remains a significant bottleneck. Many widely used corpora, such as TEACh (\citep{padmakumar2021teach}), a dataset of household interactions collected via human operation, contain a large number of suboptimal behaviors (\citep{min2022dontcopyteacher}). These include, for example, picking up irrelevant or unnecessary items, leaving goals unaccomplished due to erroneous instruction, or toggling appliances that were never activated. While such issues result in unnecessarily lengthy plans and introduce noise in learning signals during imitation or reinforcement learning, they also present an underexplored opportunity: human demonstrations naturally contain error-recovery sequences that showcase how humans recognize and correct mistakes in real-time. Our verification framework, by explicitly identifying these erroneous actions, not only enables cleaner training data through removal but also facilitates future research into learning from human error recovery patterns, potentially yielding agents that exhibit human-like resilience and adaptability when facing unexpected situations or their own execution errors.

To address these limitations, we propose a general verification framework based on a two-agent protocol. In our approach, a Planning Agent first generates a candidate plan for a given goal. This plan is then passed to a second agent – a Judge powered by an LLM – which analyzes the sequence step by step and flags actions that appear redundant, irrelevant, contradictory, or otherwise unjustified. Each flagged step is accompanied by a natural language explanation to support interpretability. The Planning Agent subsequently revises its plan in response to this feedback, and the process repeats until the Judge LLM returns no further objections or a predefined iteration limit is reached. This end-to-end sequence is described in Figure \ref{fig:multiagent-workflow}.

This approach is fully language-based, model-agnostic, and does not rely on any handcrafted heuristics or domain-specific rule sets. Instead, it leverages the reasoning capabilities of pretrained LLMs via zero-shot prompting, using a concise set of common-sense criteria embedded in natural language. The Judge LLM operates independently of any environment simulator or visual input, making the method simple to integrate into a wide range of embodied AI pipelines.

Our empirical evaluation focuses on the TEACh dataset, where we analyze 100 episodes spanning 15 high-level household tasks. We evaluate four LLMs in the Judge and Planning Agent role and report their effectiveness in identifying plan defects, measuring both precision and recall against a manually annotated ground truth. Our findings reveal a clear trade-off between conservative high-precision judges and more aggressive high-recall ones, with each offering complementary benefits for downstream applications.

\begin{figure*}[t]
\begin{center}
\includegraphics[width=14cm]{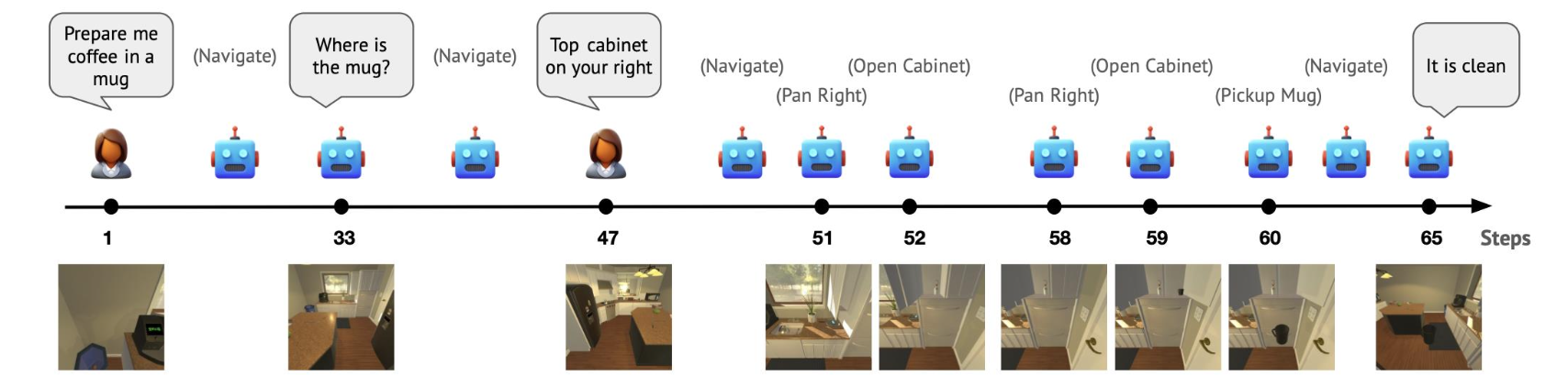} \hfill
  \caption {Diagram of Sample Workflow in TEACh Dataset}
\label{fig:teach-workflow}
\end{center}
\end{figure*}

Our contributions are:
\begin{enumerate}
\item \textbf{Framework:} We introduce a general-purpose language-based plan verification framework that incorporates iterative critique and revision into embodied planning workflows, represented in Figure \ref{fig:multiagent-workflow}.
\item \textbf{NL Prompt-Based Judge LLM:} We demonstrate that simple natural language-based rules, encoded in prompts, are sufficient to guide LLM-based judges in identifying and explaining suboptimal actions.
\item \textbf{Empirical Study:} We benchmark multiple Judge LLMs on the TEACh dataset and provide detailed analysis of their respective strengths, weaknesses, and trade-offs. The structure of these workflows is illustrated in Figure \ref{fig:teach-workflow}.
\end{enumerate}

This work lays the groundwork for scalable, language-driven refinement of human-authored task plans in embodied settings, with potential downstream benefits for both imitation learning and reinforcement learning paradigms.

\section{Related Work}
\label{sec:related}

Our work on plan verification intersects with a broad effort to integrate LLMs  into complex, multi-step reasoning pipelines. We situate our framework by synthesizing research across three major themes: the challenges and advancements in using LLMs as direct plan generators; the emergence of LLMs as verifiers and refiners in iterative feedback loops; and the broader context of the LLM-as-a-Judge ecosystem, including its benchmarks and known vulnerabilities.

\subsection{LLMs as Plan Generators}

In the earlier work, Gella et al. have demonstrated the dialogue act prediction based approach using the TEACh dataset~\cite{gella2022dialog}. This was a classification based approach trying to predict the next dialogue act given the action history and environment. They have expanded this work so as to use an LLM based planning approach in~\cite{padmakumar-etal-2023-multimodal} to generate an entire action sequence from a high-level goal. Another work by \citet{huang2022language} demonstrated that while large pre-trained models contain sufficient world knowledge to produce semantically plausible plans, these plans are often not executable because they fail to map onto an agent's specific, admissible actions. This fundamental gap between plausible text and executable actions highlights the core challenge of grounding.

Subsequent research has focused on closing this gap by providing LLMs with stronger environmental context. For instance, \textsc{ProgPrompt} \citep{singh2022progpromptgeneratingsituatedrobot} structures the prompt as a Python program, providing the LLM with available actions as import statements and objects as a list, thereby constraining the generated output to be more syntactically correct and environmentally aware. Similarly, \textsc{LLM-Planner} \citep{song2023llmplanner} couples few-shot prompting with online simulator feedback, appending error traces to the prompt to enable replanning when a step fails. Other methods, like \textsc{TaPA} \citep{wu2023tapa}, inject explicit world state, such as an inventory table, directly into the prompt to reduce object hallucination. Our work is orthogonal to these plan generation methods; instead of improving the initial output, we focus on creating a robust post-hoc verification process to clean noisy, human-authored trajectories, ensuring that any downstream planner inherits a leaner and less ambiguous set of demonstrations.

\subsection{Verification and Refinement}

Given the difficulties in direct plan generation, a powerful alternative paradigm has emerged: using LLMs to critique, verify, and refine plans within an iterative loop. This approach often decomposes the problem into distinct agentic roles. The Reflexion framework, for example, splits an agent into an Actor, Evaluator, and Self-Reflection module, using verbal reinforcement to iteratively improve its policy without gradient updates \citep{shinn2023reflexion}. Our Judge-Planner framework externalizes this critique process into an independent, two-agent system, which facilitates explicit precision-recall accounting and allows for modular substitution of the Judge LLM. This multi-agent perspective is echoed in broader frameworks like \textsc{AutoGen}, which coordinates specialized conversable agents \citep{wu2023autogen}, and \textsc{CoELA}, which pairs a "Commander" LLM with domain-specific modules for robot teams \citep{zhang2023coela}.

More directly, recent work has built systems specifically for pre-execution plan verification. VerifyLLM \citep{grigorev2025verifyllm} proposes a verification architecture that analyzes action sequences to identify logical inconsistencies, missing prerequisites, and redundancies, using Linear Temporal Logic (LTL) as a formal intermediate representation to guide the LLM's analysis. The principle of using an LLM to verify an intermediate output is not limited to robotics; LLatrieval \citep{li2024llatrieval} employs an LLM to iteratively verify and refine the documents retrieved by a RAG system, demonstrating the general power of this verification-and-update pattern. Our framework contributes to this line of work by demonstrating that effective, high-recall verification can be achieved using zero-shot natural language critique alone, without requiring formal methods like LTL.

\subsection{"LLM-as-a-Judge" Ecosystem}

Casting an LLM in a verification role makes it an instance of an "LLM-as-a-Judge," a rapidly growing field with its own set of tools and challenges. A systematic analysis by \citet{li2024systematic} formally evaluates the distinct contributions of LLMs as solvers, verifiers, and heuristic functions, concluding that LLMs are often more effective at providing comparative feedback than at generating correct solutions from scratch – a finding that strongly motivates our verification-centric approach.

However, the reliability of LLM judges is not guaranteed. Research into their adversarial robustness has shown that simple, universal triggers can inflate judge scores, highlighting the fragility of naive evaluation \citep{raina2024robust}. To promote standardized evaluation, public benchmarks such as \textsc{MT-Bench} and the \textsc{Chatbot Arena} have been developed to provide open, crowd-validated tests where models like GPT-4 have achieved human-level agreement \citep{zheng2023mtbench}. Our work adopts their strict evaluation protocols. Furthermore, systematic surveys have begun to chart the field, offering taxonomies of judgment tasks and techniques for bias mitigation \citep{gu2025survey, li2025generation}, which inform the design of our evaluation suite. Finally, within the embodied domain, benchmarks like \textsc{PARTNR} provide large-scale, multi-agent tasks with built-in verification checkpoints \citep{chang2022partnr}. We instead target the TEACh dataset for its organic, noisy human demonstrations, which better approximate the data our framework is designed to clean.


\section{Methodology}
\label{sec:method}

Our objective is to automatically refine \emph{human-authored} embodied task trajectories by identifying and correcting actions that are irrelevant, redundant, contradictory, or missing relative to the stated goal. We cast this process as an interaction between two language model-based agents: a \textbf{Judge LLM}, which critiques a proposed plan, and a \textbf{Planning Agent}, which applies those critiques to produce an updated plan. This modular design yields a transparent, fully language-driven verification loop that can be applied to existing datasets without retraining.

\subsection{Formal Problem Definition}

Let $\Pi$ denote the space of action sequences (plans). A plan $\pi \in \Pi$ is a finite sequence
\[
\pi = (a_1, a_2, \dots, a_T),
\]
where each $a_t$ is an atomic manipulation such as \texttt{PickUp}, \texttt{ToggleOn}, or \texttt{Place}. Not all actions in $\pi$ are correct. We consider an \emph{error set} $\mathcal{E}(\pi)$ containing the positions of erroneous actions in $\pi$. Errors arise in three forms:
\begin{enumerate}
    \item \textbf{Redundant actions} (unnecessary repetitions or canceling operations).
    \item \textbf{Contradictory actions} (steps that oppose earlier actions or the task goal; treated as \texttt{REMOVE}).
    \item \textbf{Missing actions} (omissions required for successful task completion).
\end{enumerate}

\paragraph{Objective.}  
Given a natural language goal $g$ and an initial plan $\pi^{(0)}$, our task is to produce a refined plan $\pi^*$ that achieves $g$ while minimising length:
\[
\pi^* = \arg\min_{\tilde \pi}\;|\tilde \pi| \quad \text{s.t.}\quad \tilde \pi \text{ achieves } g.
\]
This formulation permits insertions when necessary, but penalises unnecessary actions.

\paragraph{Verification Operator.}  
The Judge implements a critique function
\[
J : (g,\pi) \mapsto \mathcal{C} = \{(i, \text{type}, \text{reason})\},
\]
where $\text{type}\in\{\texttt{REMOVE},\texttt{MISSING}\}$ and each critique specifies an action index $i$, a correction type, and a rationale. The Planner applies these critiques deterministically:
\[
P : (\pi,\mathcal{C}) \mapsto \pi'.
\]
We define the verification operator as the composition
\[
V = P \circ J,
\]
so that $\pi' = V(g,\pi)$. We assume $V$ is \emph{conservative}, i.e.
\[
\mathbb{E}[E(V(\pi))] \leq E(\pi),
\]
where $E(\pi) = |\mathcal{E}(\pi)|$ is the error count.

\paragraph{Iterative Refinement.}  
Verification is applied iteratively. At iteration $k$, the Judge produces critiques $\mathcal{C}^{(k)} = J(g,\pi^{(k)})$, and the Planner incorporates them:
\[
\pi^{(k+1)} = P(\pi^{(k)},\mathcal{C}^{(k)}).
\]
The error counts form a non-increasing sequence $\{E^{(k)}\}$. Under the conservative assumption, there exists $\delta>0$ such that
\[
\mathbb{E}[E^{(k+1)}] \leq (1-\delta)\,\mathbb{E}[E^{(k)}],
\]
suggesting geometric convergence to a fixed point $E^* \geq 0$. Empirically, convergence occurs within three iterations in 96.5\% of cases.

\subsection{Evaluation Metrics}

We evaluate the Judge’s ability to detect errors against human annotations.  
\begin{itemize}
    \item \textbf{Recall} = $\tfrac{\text{TP}}{\text{TP}+\text{FN}}$: fraction of erroneous actions flagged.  
    \item \textbf{Precision} = $\tfrac{\text{TP}}{\text{TP}+\text{FP}}$: fraction of flagged actions that were truly erroneous.  
    \item \textbf{F$_1$ score}: harmonic mean of precision and recall.  
\end{itemize}
Here positives are erroneous actions (\textsc{Remove} or \textsc{Missing}); contradictory errors are subsumed under \textsc{Remove}.  
Task-level outcomes (e.g., plan success rate, average length reduction) are reported separately.

\subsection{Iterative Verification Algorithm}

Algorithm~\ref{alg:plan_verification} instantiates the operator $V = P \circ J$ as an iterative refinement loop.

\begin{algorithm}[h]
\caption{Iterative Plan Verification (Judge–Planner Composition)}
\label{alg:plan_verification}
\begin{algorithmic}
\Require Goal $g$, initial plan $A = (a_1, \ldots, a_n)$, Judge LLM $J$, Planner LLM $P$
\Ensure Refined plan $A'$
\State $A' \gets A$
\For{$i = 1$ to $5$} \Comment{Maximum refinement rounds}
    \State $\mathit{critiques} \gets J.\text{evaluate}(g, A')$
    \If{$\mathit{critiques} = \emptyset$}
        \State \Return $A'$ \Comment{No errors detected}
    \EndIf
    \State $A' \gets P.\text{apply\_critiques}(A', \mathit{critiques})$
    \State $A' \gets \text{reindex}(A')$
\EndFor
\State \Return $A'$ \Comment{Return last refined plan if critiques persist}
\end{algorithmic}
\end{algorithm}


\section{Results}
\label{sec:results}

We report results from both the single-pass (zero-shot) and multi-pass (iterative) evaluation settings. These results are obtained by manual annotations that assess the final Planning Agent output based on the action sequence accurately achieving the goal set by the Commander of each TEACh action sequence.

\subsection{Static (Zero-Shot) Verification Performance}

Table~\ref{tab:static_results} summarizes the recall and precision of each Judge LLM when applied directly to raw TEACh plans. GPT o4-mini achieved the highest overall recall (80\%) while maintaining a strong precision of 93\%. DeepSeek-R1 exhibited perfect precision (100\%) but with a lower recall of 68\%, indicating a more conservative judging style. Gemini 2.5 and LLaMA 4 Scout provided moderately strong recall (74\%) with slightly reduced precision (90\% and 85\%, respectively). The rule-based baseline underperformed across both metrics, as shown in Table~\ref{tab:static_results}.

\begin{table}[ht]
\centering
\begin{tabular}{lcc}
\toprule
\textbf{Judge LLM} & \textbf{Recall} & \textbf{Precision} \\
\midrule
GPT o4-mini & \textbf{80\%} & 93\% \\
DeepSeek-R1 & 68\% & \textbf{100\%} \\
Gemini 2.5 & 74\% & 90\% \\
LLaMA 4 Scout & 74\% & 85\% \\
Rule-based & 22\% & 71\% \\
\bottomrule
\end{tabular}
\caption{Single-pass plan verification performance.}
\label{tab:static_results}
\end{table}

\subsection{Iterative Critique-and-Revise Performance}

We now examine how plan quality improves over multiple critique–and-revise rounds. Each Planning Agent LLM proposes a revised plan after receiving feedback from the Judge LLM, and this loop continues until the Judge LLM raises no new issues.

As shown in Table~\ref{tab:iterative_recall}, iterative refinement boosts recall by 5–10\% on average across all Judge–Planner combinations. For instance, GPT o4-mini’s recall improves from 80\% (static) to 88–90\% when paired with itself or other Planner LLMs. Similarly, Gemini 2.5 sees recall rise to 89\% with multiple planner pairings. Precision generally remains stable or improves slightly. 

\begin{figure}[h]
\centering
\includegraphics[width=0.65\textwidth]{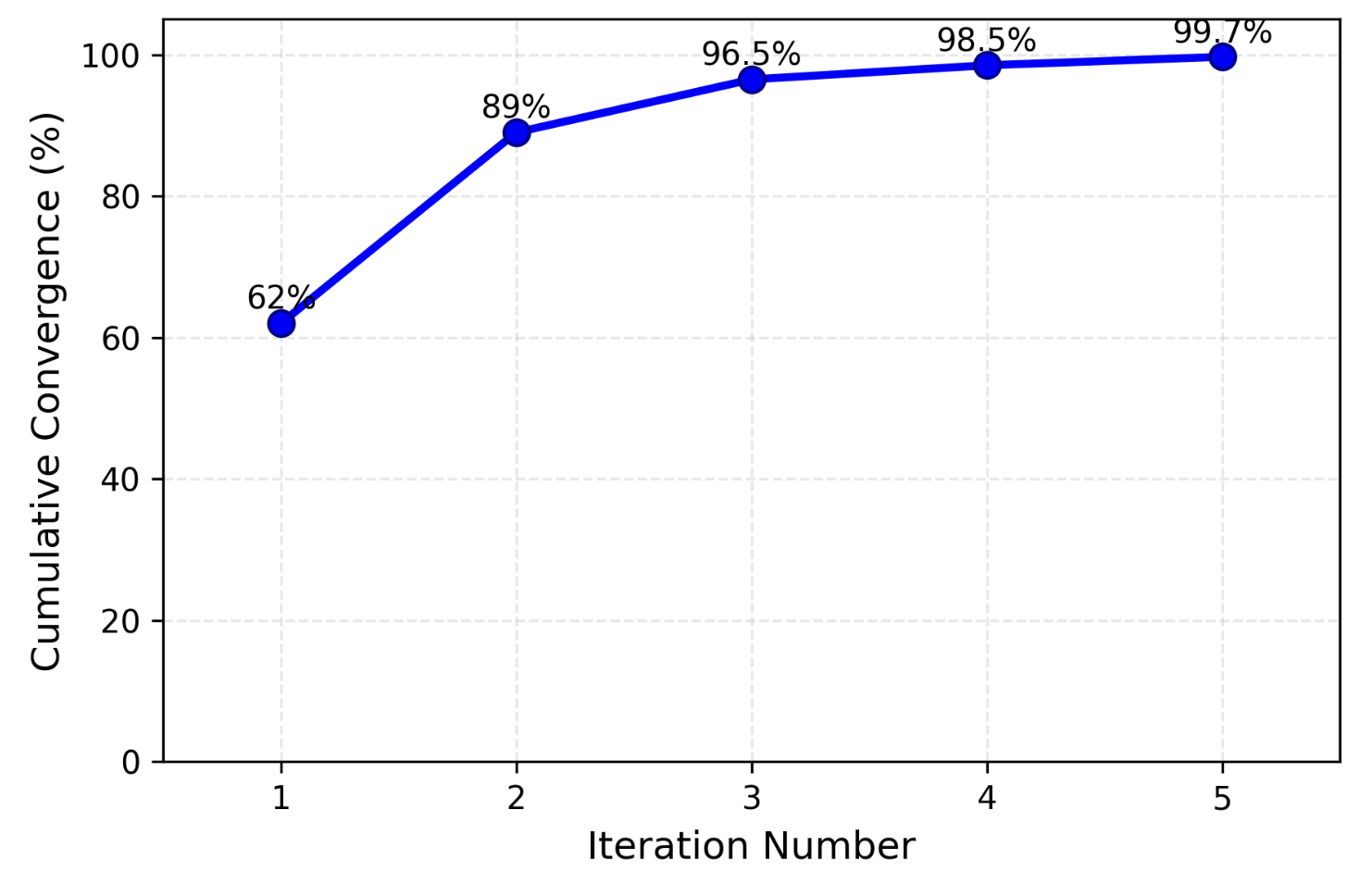}
\caption{Cumulative convergence of action sequences across iterations. Most sequences (62\%) are corrected after the first iteration, with near-complete convergence (96.5\%) by iteration 3.}
\label{fig:convergence}
\end{figure}

To understand the efficiency of the iterative refinement process, we analyzed the convergence behavior across all each of the individual steps over all actions cumulatively judged and planned by the LLM agents. Figure~\ref{fig:convergence} illustrates the cumulative percentage of sequences that reach their final state after each iteration.

Our analysis reveals:
\begin{itemize}
\itemsep -0.5ex
    \item \textbf{Iteration 1}: 62\% of sequences require no further modifications
    \item \textbf{Iteration 2}: 89\% cumulative convergence (+27\%)
    \item \textbf{Iteration 3}: 96.5\% cumulative convergence (+7.5\%)
    \item \textbf{Iterations 4-5}: Only 3.5\% of sequences benefit from additional rounds
\end{itemize}

This rapid convergence suggests that most plan defects are straightforward and can be identified in a single pass, while a smaller subset requires iterative reasoning to resolve complex interdependencies between actions.

\begin{table*}[ht]
\centering
\begin{tabular}{lcccc}
\toprule
\textbf{Judge LLM} & \multicolumn{4}{c}{\textbf{Planner LLM – Recall (\%) / Precision (\%) / F-score}} \\
\cmidrule{2-5}
 & GPT o4-mini & DeepSeek-R1 & Gemini 2.5 & LLaMA 4 Scout \\
\midrule
GPT o4-mini & 88 / 90 / 89.0 & \textbf{90 / 80 / 84.7} & 85 / 91 / 87.8 & 89 / 87 / 87.9 \\
DeepSeek-R1 & 65 / 99 / 78.5 & \textbf{68 / 100 / 80.9} & 62 / 100 / 76.5 & 66 / 98 / 78.9 \\
Gemini 2.5 & 84 / 98 / 90.7 & 86 / 97 / 91.2 & \textbf{89 / 99 / 93.9} & 89 / 96 / 92.2 \\
LLaMA 4 Scout & 76 / 92 / 83.5 & 81 / 90 / 85.3 & 79 / 93 / 85.9 & 75 / 89 / 81.6 \\
\bottomrule
\end{tabular}
\caption{Iterative recall and precision (\%): Recall is the percentage of ground-truth \textsc{Remove}/\textsc{Missing} actions correctly flagged; precision is the percentage of flagged actions that were genuinely erroneous}
\label{tab:iterative_recall}
\end{table*}

\subsection{Qualitative Analysis and Error Patterns}

Our manual analysis of Judge LLM outputs reveals systematic patterns in both successful corrections and failure modes. We categorize these findings by error type and precision/recall failures.

\subsubsection{Model Behavior}

Our analysis reveals distinct behavioral patterns among the four Judge LLMs evaluated. \textbf{DeepSeek-R1} emerges as the most conservative judge, achieving near-perfect precision (98-100\%) across all planner pairings but at the cost of significantly lower recall (62-68\%). This conservative approach means DeepSeek-R1 only flags actions when absolutely certain, resulting in many problematic actions passing through undetected. While this minimizes false positives, it fails to identify approximately one-third of the errors in human-authored plans, limiting its effectiveness for comprehensive plan refinement.

In contrast, \textbf{GPT o4-mini} demonstrates a more balanced but aggressive approach, achieving the highest recall rates (85-90\%) while maintaining respectable precision (80-91\%). This model excels at catching subtle errors that other judges miss, particularly when paired with DeepSeek-R1 as the planner (90\% recall). However, its eagerness to flag potential issues occasionally leads to over-correction, with precision dropping to 80\% in some configurations. This trade-off makes GPT o4-mini ideal for scenarios where catching all errors is paramount, even if some valid actions are incorrectly flagged.

\textbf{Gemini 2.5} presents the most well-rounded performance profile, consistently achieving high F1-scores (90.7-93.9) that indicate excellent balance between precision and recall. With recall rates of 84-89\% and precision consistently above 96\%, Gemini 2.5 demonstrates sophisticated judgment capabilities that avoid both the over-conservatism of DeepSeek-R1 and the occasional over-eagerness of GPT o4-mini. Notably, Gemini 2.5 performs best when paired with itself or LLaMA 4 Scout as planners, suggesting strong internal consistency in its evaluation criteria.

\textbf{LLaMA 4 Scout} occupies a middle ground, with moderate recall (75-81\%) and good precision (89-93\%). While it doesn't excel in any particular metric, its consistent performance across different planner pairings suggests robustness to varying input styles. LLaMA 4 Scout appears to focus on identifying the most obvious errors while being cautious about borderline cases, making it a reliable if unspectacular choice for plan verification tasks.

The interaction effects between judge-planner pairs also reveal interesting patterns. Models generally perform best when paired with themselves (as seen in the diagonal of our results table), suggesting that each model has internally consistent standards for what constitutes correct action sequences. However, some cross-model pairings yield surprising benefits: GPT o4-mini as judge with DeepSeek-R1 as planner achieves the highest recall (90\%), while Gemini 2.5 consistently achieves the best overall performance regardless of planner choice, indicating its superior generalization capabilities.

\subsubsection{Successful Corrections}

Judge LLMs consistently identify and correct:
\begin{itemize}
\item \textbf{Premature toggles}: \texttt{REMOVE: Driver toggles off microwave before turning it on}
\item \textbf{Irrelevant objects}: \texttt{REMOVE: Driver picks up RemoteControl – not needed for cooking task}
\item \textbf{Incomplete sequences}: \texttt{MISSING: Task incomplete – bread sliced but sandwich not assembled}
\end{itemize}

\subsubsection{Recall Failures (Missed Errors)}

The Judge LLMs systematically miss certain error patterns, which are marked by the annotator's notes (AN) in the examples below:

\paragraph{Context-dependent redundancies.} In sequences where objects are picked up early but used much later, judges fail to recognize the inefficiency:
\begin{verbatim}
// GOAL: Clean the bathroom
Driver.PickUp('Soap')      // Picked up early
Driver.Move(5.0)           // Multiple intervening actions
Driver.Turn(90)            
Driver.PickUp('Sponge')    
Driver.Place('Sink')       // Soap finally used here
// AN: Judge fails to flag early pickup as inefficient
\end{verbatim}

\subsubsection{Precision Failures (False Positives)}

Judge LLMs occasionally over-correct valid sequences:

\paragraph{Multi-step preparations.} Judges sometimes flag necessary preparatory actions as redundant:
\begin{verbatim}
// GOAL: Make coffee
Driver.PickUp('Mug')        // REMOVE (AN: Incorrectly flagged)
Driver.Place('Counter')
Driver.PickUp('CoffeeFilter')
Driver.Place('CoffeeMachine')
\end{verbatim}

\paragraph{Object reuse patterns.} When objects serve multiple purposes, judges may incorrectly remove secondary uses:
\begin{verbatim}
// GOAL: Set dinner table
Driver.PickUp('Plate')
Driver.Place('DiningTable')
Driver.PickUp('Plate')      // REMOVE (AN: Incorrectly flagged)
Driver.Place('DiningTable') // Actually placing second plate
\end{verbatim}

These error patterns suggest that while current Judge LLMs excel at surface-level logical inconsistencies, they may struggle with long-range dependencies and context-sensitive reasoning.


\section{Conclusion}

This paper presents a general-purpose, language-driven framework for verifying human-authored embodied task plans via structured LLM-based critique. By casting plan verification as a modular dialogue between a Planning Agent and a Judge LLM, we enable scalable, interpretable, and model-agnostic refinement of noisy demonstration datasets without the need for task-specific heuristics or environment simulators. We evaluated the four LLMs GPT o4-mini, DeepSeek-R1, Gemini 2.5, and LLaMA 4 Scout as Judges on a curated slice of the TEACh dataset comprising 1,408 human-generated actions. Our findings reveal consistent advantages of natural language-based verification over rule-based baselines, with zero-shot prompting alone yielding up to 80\% recall and 100\% precision depending on the model. Moreover, we demonstrate that iterative critique-and-revise loops provide a consistent 5–10\% boost in recall without sacrificing precision, underscoring the complementary power of multi-round plan inspection and LLM collaboration. Qualitative analysis confirms that Judge LLMs produce human-interpretable, step-level rationales that support transparent auditability – providing a more interpretable alternative. We also identify trade-offs between precision-focused and recall-focused Judge LLMs, suggesting hybrid ensembles or confidence-calibrated prompts as promising future directions.

These results establish plan verification as a distinct, language-level capability of LLMs, with strong implications for both dataset curation and agent performance in downstream learning tasks. Future work will extend this framework to vision-grounded settings, explore adversarial robustness of judge prompts, and test the framework’s scalability on larger multi-agent corpora such as PARTNR~\citep{chang2022partnr}.


\section{Limitations}

Our study introduces a robust verification framework utilizing natural language-based critique-and-revision loops to refine action plans within embodied task scenarios. However, several limitations must be acknowledged. Firstly, our evaluation is conducted exclusively on a subset of the TEACh dataset, covering 15 common household goals, and the generalizability of our findings to broader, more diverse embodied tasks and environments (e.g., industrial settings, healthcare, or complex outdoor environments) remains untested. Further, our results depend significantly on manual labeling of actions as necessary or unnecessary, and although careful and systematic, this process inherently introduces subjective judgment and potential inconsistencies, suggesting that future research should explore automated or semi-automated annotation methods to enhance consistency and efficiency.

The efficacy of our approach hinges on the quality and reliability of the Judge LLMs used as well, and the zero-shot prompting method, while effective, may be limited by the LLM's inherent biases, knowledge cutoff dates, and potential hallucinations or inaccuracies. These issues could potentially be mitigated through fine-tuning or hybrid approaches integrating external knowledge bases. Moreover, while our iterative critique-and-revise loop is computationally lightweight once the Judge LLM responses are cached, the initial computational overhead of generating Judge LLM responses, especially with large-scale datasets or more complex tasks, could be considerable, necessitating optimizations or more efficient prompting strategies for scalability. Finally, our method currently relies solely on natural language cues for identifying redundant or erroneous actions, and integrating stronger environmental grounding (e.g., visual object recognition or physical simulations) may enhance the accuracy and reliability of the verification process. Addressing these limitations through further research will be essential to fully realize the potential of natural-language verification approaches in diverse and complex real-world embodied tasks.

\section{Acknowledgements}
This research was supported in part by Other Transaction award HR0011249XXX from the U.S. Defense Advanced Research Projects Agency (DARPA) Friction for Accountability in Conversational Transactions (FACT) program

\bibliographystyle{unsrtnat}
\bibliography{custom}

@inproceedings{shinn2023reflexion,
      title={Reflexion: Language Agents with Verbal Reinforcement Learning}, 
      author={Noah Shinn and Federico Cassano and Edward Berman and Ashwin Gopinath and Karthik Narasimhan and Shunyu Yao},
      year={2023},
booktitle = {Proceedings of the 37th International Conference on Neural Information Processing Systems}, articleno = {377}, numpages = {19},
      eprint={2303.11366},
      archivePrefix={arXiv},
      primaryClass={cs.AI},
      url={https://arxiv.org/abs/2303.11366}, 
}

@article{wu2023autogen,
      title={AutoGen: Enabling Next-Gen LLM Applications via Multi-Agent Conversation}, 
      author={Qingyun Wu and Gagan Bansal and Jieyu Zhang and Yiran Wu and Beibin Li and Erkang Zhu and Li Jiang and Xiaoyun Zhang and Shaokun Zhang and Jiale Liu and Ahmed Hassan Awadallah and Ryen W White and Doug Burger and Chi Wang},
      year={2023},
      eprint={2308.08155},
      archivePrefix={arXiv},
      primaryClass={cs.AI},
      url={https://arxiv.org/abs/2308.08155}, 
}

@article{padmakumar2021teach,
      title={TEACh: Task-driven Embodied Agents that Chat}, 
      author={Aishwarya Padmakumar and Jesse Thomason and Ayush Shrivastava and Patrick Lange and Anjali Narayan-Chen and Spandana Gella and Robinson Piramuthu and Gokhan Tur and Dilek Hakkani-Tur},
      year={2021},
      eprint={2110.00534},
      archivePrefix={arXiv},
      primaryClass={cs.CV},
      url={https://arxiv.org/abs/2110.00534}, 
}

@article{min2022dontcopyteacher,
      title={Don't Copy the Teacher: Data and Model Challenges in Embodied Dialogue}, 
      author={So Yeon Min and Hao Zhu and Ruslan Salakhutdinov and Yonatan Bisk},
      year={2022},
      eprint={2210.04443},
      archivePrefix={arXiv},
      primaryClass={cs.LG},
      url={https://arxiv.org/abs/2210.04443}, 
}

@inproceedings{padmakumar-etal-2023-multimodal,
    title = "Multimodal Embodied Plan Prediction Augmented with Synthetic Embodied Dialogue",
    author = "Padmakumar, Aishwarya  and
      Inan, Mert  and
      Gella, Spandana  and
      Lange, Patrick  and
      Hakkani-Tur, Dilek",
    editor = "Bouamor, Houda  and
      Pino, Juan  and
      Bali, Kalika",
    booktitle = "Proceedings of the 2023 Conference on Empirical Methods in Natural Language Processing",
    month = dec,
    year = "2023",
    address = "Singapore",
    publisher = "Association for Computational Linguistics",
    url = "https://aclanthology.org/2023.emnlp-main.374/",
    doi = "10.18653/v1/2023.emnlp-main.374",
    pages = "6114--6131"
}

@article{gella2022dialog,
      title={Dialog Acts for Task-Driven Embodied Agents}, 
      author={Spandana Gella and Aishwarya Padmakumar and Patrick Lange and Dilek Hakkani-Tur},
      year={2022},
      eprint={2209.12953},
      archivePrefix={arXiv},
      primaryClass={cs.CL},
      url={https://arxiv.org/abs/2209.12953}, 
}

@article{zhang2023coela,
      title={Building Cooperative Embodied Agents Modularly with Large Language Models}, 
      author={Hongxin Zhang and Weihua Du and Jiaming Shan and Qinhong Zhou and Yilun Du and Joshua B. Tenenbaum and Tianmin Shu and Chuang Gan},
      year={2024},
      eprint={2307.02485},
      archivePrefix={arXiv},
      primaryClass={cs.AI},
      url={https://arxiv.org/abs/2307.02485}, 
}

@misc{song2023llmplanner,
      title={LLM-Planner: Few-Shot Grounded Planning for Embodied Agents with Large Language Models}, 
      author={Chan Hee Song and Jiaman Wu and Clayton Washington and Brian M. Sadler and Wei-Lun Chao and Yu Su},
      year={2023},
      eprint={2212.04088},
      archivePrefix={arXiv},
      primaryClass={cs.AI},
      url={https://arxiv.org/abs/2212.04088}, 
}

@misc{wu2023tapa,
      title={Embodied Task Planning with Large Language Models}, 
      author={Zhenyu Wu and Ziwei Wang and Xiuwei Xu and Jiwen Lu and Haibin Yan},
      year={2023},
      eprint={2307.01848},
      archivePrefix={arXiv},
      primaryClass={cs.CV},
      url={https://arxiv.org/abs/2307.01848}, 
}

@article{chang2022partnr,
  title={PARTNR: A Benchmark for Planning and Reasoning in Embodied Multi-agent Tasks},
  author={Chang, Matthew and Chhablani, Gunjan and Clegg, Alexander and Cote, Mikael Dallaire and Desai, Ruta and Hlavac, Michal and Karashchuk, Vladimir and Krantz, Jacob and Mottaghi, Roozbeh and Parashar, Priyam and Patki, Siddharth and Prasad, Ishita and Puig, Xavier and Rai, Akshara and Ramrakhya, Ram and Tran, Daniel and Truong, Joanne and Turner, John M. and Undersander, Eric and Yang, Tsung-Yen},
  journal={arXiv preprint arXiv:2205.08866},
  year={2022},
  note={Work done at FAIR Meta. Alphabetical author order}
}

@article{raina2024robust,
      title={Is LLM-as-a-Judge Robust? Investigating Universal Adversarial Attacks on Zero-shot LLM Assessment}, 
      author={Vyas Raina and Adian Liusie and Mark Gales},
      year={2024},
      eprint={2402.14016},
      archivePrefix={arXiv},
      primaryClass={cs.CL},
      url={https://arxiv.org/abs/2402.14016}, 
}

@misc{zheng2023mtbench,
      title={Judging LLM-as-a-Judge with MT-Bench and Chatbot Arena}, 
      author={Lianmin Zheng and Wei-Lin Chiang and Ying Sheng and Siyuan Zhuang and Zhanghao Wu and Yonghao Zhuang and Zi Lin and Zhuohan Li and Dacheng Li and Eric P. Xing and Hao Zhang and Joseph E. Gonzalez and Ion Stoica},
      year={2023},
      eprint={2306.05685},
      archivePrefix={arXiv},
      primaryClass={cs.CL},
      url={https://arxiv.org/abs/2306.05685}, 
}

@misc{gu2025survey,
      title={A Survey on LLM-as-a-Judge}, 
      author={Jiawei Gu and Xuhui Jiang and Zhichao Shi and Hexiang Tan and Xuehao Zhai and Chengjin Xu and Wei Li and Yinghan Shen and Shengjie Ma and Honghao Liu and Saizhuo Wang and Kun Zhang and Yuanzhuo Wang and Wen Gao and Lionel Ni and Jian Guo},
      year={2025},
      eprint={2411.15594},
      archivePrefix={arXiv},
      primaryClass={cs.CL},
      url={https://arxiv.org/abs/2411.15594}, 
}

@article{li2025generation,
      title={From Generation to Judgment: Opportunities and Challenges of LLM-as-a-judge}, 
      author={Dawei Li and Bohan Jiang and Liangjie Huang and Alimohammad Beigi and Chengshuai Zhao and Zhen Tan and Amrita Bhattacharjee and Yuxuan Jiang and Canyu Chen and Tianhao Wu and Kai Shu and Lu Cheng and Huan Liu},
      year={2025},
      eprint={2411.16594},
      archivePrefix={arXiv},
      primaryClass={cs.AI},
      url={https://arxiv.org/abs/2411.16594}, 
}

@article{li2024llatrieval,
      title={LLatrieval: LLM-Verified Retrieval for Verifiable Generation}, 
      author={Xiaonan Li and Changtai Zhu and Linyang Li and Zhangyue Yin and Tianxiang Sun and Xipeng Qiu},
      year={2024},
      eprint={2311.07838},
      archivePrefix={arXiv},
      primaryClass={cs.CL},
      url={https://arxiv.org/abs/2311.07838}, 
}

@article{huang2022language,
      title={Language Models as Zero-Shot Planners: Extracting Actionable Knowledge for Embodied Agents}, 
      author={Wenlong Huang and Pieter Abbeel and Deepak Pathak and Igor Mordatch},
      year={2022},
      eprint={2201.07207},
      archivePrefix={arXiv},
      primaryClass={cs.LG},
      url={https://arxiv.org/abs/2201.07207}, 
}

@article{singh2022progpromptgeneratingsituatedrobot,
      title={ProgPrompt: Generating Situated Robot Task Plans using Large Language Models}, 
      author={Ishika Singh and Valts Blukis and Arsalan Mousavian and Ankit Goyal and Danfei Xu and Jonathan Tremblay and Dieter Fox and Jesse Thomason and Animesh Garg},
      year={2022},
      eprint={2209.11302},
      archivePrefix={arXiv},
      primaryClass={cs.RO},
      url={https://arxiv.org/abs/2209.11302}, 
}

@article{li2024systematic,
      title={Systematic Analysis of LLM Contributions to Planning: Solver, Verifier, Heuristic}, 
      author={Haoming Li and Zhaoliang Chen and Songyuan Liu and Yiming Lu and Fei Liu},
      year={2024},
      eprint={2412.09666},
      archivePrefix={arXiv},
      primaryClass={cs.AI},
      url={https://arxiv.org/abs/2412.09666}, 
}

@article{grigorev2025verifyllm,
      title={VerifyLLM: LLM-Based Pre-Execution Task Plan Verification for Robots}, 
      author={Danil S. Grigorev and Alexey K. Kovalev and Aleksandr I. Panov},
      year={2025},
      eprint={2507.05118},
      archivePrefix={arXiv},
      primaryClass={cs.RO},
      url={https://arxiv.org/abs/2507.05118}, 
}

\appendix

\section{Judge LLM and Planning Agent Prompts}
\label{app:judge-prompt}

The following prompt was used to instruct the LLMs to evaluate and flag unnecessary or erroneous actions. 

\subsection{Judge LLM Prompt}
\begin{verbatim}
You are a Judge Agent for embodied AI task planning. Your role is to provide
 thoughtful,natural language feedback on action sequences. You should:
1. Analyze each action's purpose and relevance to the goal
2. Explain your reasoning in clear, conversational language
3. Point out redundant or unnecessary actions with detailed explanations
4. Identify missing actions needed to complete the goal
5. Focus on being helpful and constructive in your feedback
Provide your feedback as natural language commentary, using #REMOVE and 
#MISSING tags only when necessary. Prioritize clear explanations.
Please evaluate this action sequence for achieving the following goal:
GOAL: {goal}
Action Sequence: {actions_text}
Provide line-by-line analysis of each action. For each action, explain what
it does and whether it's necessary for the goal. Use this format:
ACTION: [copy the exact action]
ANNOTATION: [explain what this action does and whether it's needed for the goal. 
If the action should be removed, include "#REMOVE: reason".
If it's good, just explain why.]
After analyzing all actions, if any steps are missing to complete the goal, add:
#MISSING: [describe what actions are needed]
Be thorough and conversational in your explanations. Focus on helping someone 
understand why each action is or isn't necessary for achieving the goal.
Your line-by-line analysis:
\end{verbatim}

\subsection{Planning Agent Prompt}
\begin{verbatim}
You are a Planning Agent for embodied AI tasks. Your role is to:
1. Analyze action sequences and identify their goals
2. Modify action sequences based on feedback from a Judge
3. Remove redundant actions and add missing actions as needed
4. Ensure action sequences are efficient and complete
Always preserve the original format and only make necessary changes.
Analyze the following action sequence and determine the overall 
Context: {context}
Actions:{actions_text}
Provide a concise goal statement starting with "GOAL: "
\end{verbatim}

\end{document}